# Faster method for Deep Belief Network based Object Classification using DWT

Saurabh Sihag and Pranab Kr. Dutta

A Deep Belief Network (DBN) requires large, multiple hidden layers with high number of hidden units to learn good features from the raw pixels of large images. This implies more training time as well as computational complexity. By integrating DBN with Discrete Wavelet Transform (DWT), both training time and computational complexity can be reduced. The low resolution images obtained after application of DWT are used to train multiple DBNs. The results obtained from these DBNs are combined using a weighted voting algorithm. The performance of this method is found to be competent and faster in comparison with that of traditional DBNs.

*Introduction:* Deep learning has found applications in various areas of information processing such as audio processing, natural language modeling and processing, object recognition and computer vision because of the high accuracy of its models. Deep Belief Networks (DBNs) learn complex function mapping from input to output directly from raw pixels of data. DBN training, which includes the pre-training and fine-tuning processes, in conventional central processing units (CPU) platforms is computationally expensive because multiple hidden layers with high number of hidden units are required to train the mapping function for high dimensional raw pixel data. Large amounts of training data are needed to learn the parameters of such a network for preventing overfitting. Execution of Deep Learning algorithms can be made faster by reducing the dimensionality of the data, as illustrated in recent years. One such approach is presented in [1] which illustrates the use of Discrete Cosine Transform (DCT) for image classification. Wavelet transform, rough set theory, and artificial neural networks are combined together to form a hybrid image classification method in [2]. Multiresolution image features have been utilized for object detection in [3].

In this letter, a new approach is proposed for image classification that integrates discrete wavelet transform (DWT) and DBNs. The idea is to break up an input image into several low resolution sub-band images using 2 level DWT and train a separate DBN for each sub-band image. These DBNs use less number of hidden units and require less training time and memory individually as compared to that of a traditional DBN that is trained using raw pixels from the input data. The classification results obtained from each DBN are processed using a weighted voting algorithm to achieve the final result.

*Theoretical Description:* Restricted Boltzmann machines (RBMs) [4] are probabilistic models that are used as nonlinear unsupervised feature learners, consisting of a set of binary hidden units $h$, a set of (binary or real-valued) visible units $v$, and a weight matrix $W$ associated with the connections between the two layers. Joint probability function for visible and hidden units is defined as

$$P(v,h) = \frac{1}{Z}\exp(-E(v,h))$$

where $Z$ is the partition function and $E(v,h)$ is the energy function. For an RBM with real valued visible units, the energy function is defined as:

$$E(v,h) = -\sum_{s=1}^{n^v} b_s^v v_s - \sum_{t=1}^{n^h} b_t^h h_t - \sum_{s=1}^{n^v}\sum_{t=1}^{n^h} v_s h_t w_{st}$$

where $v_s$ and $h_t$ are the $s^{th}$ and $t^{th}$ units of $v$ and $h$, $b_s^v$ and $b_t^h$ are the biases associated with unit $v_s$ and $h_t$, and $w_{st}$ is the weight associated with the connection between $v_s$ and $h_t$. A deep belief network (DBN) consists of layers of RBMs, each of which contains a layer of visible units and a layer of hidden units. Two adjacent layers have a full set of connections between them, but no two units in the same layer are connected. An efficient algorithm for training deep belief networks has been proposed in [5]. Discrete Wavelet Transform (DWT) is used to decompose an original image into different sub-band images, with each sub-band image occupying low and high frequency bands.

*Methodology:*
*A. Generating low resolution images from the input image*
Two level DWT is used to get sixteen sub-band images from the input image. Unlike the conventional DWT process, all the four sub-band images obtained after first level of DWT are decomposed further to generate four sub-band images each or a total of sixteen sub-band images. Size of each image is one-sixteenth of the size of original image.

*B. Formation of Deep Belief Networks*
The sixteen low resolution images obtained from the previous step are used as the training features for DBNs. Sixteen DBNs are formed corresponding to each low resolution image obtained from the previous step. Each DBN has same number of hidden layers as well as hidden units.

*C. Training the DBNs*
Initially each DBN is pre-trained using Contrastive Divergence, as specified in [5]. This is the unsupervised training step. In the second step, back Propagation is used for fine-tuning of the weights.
The pre-training step helps to achieve faster convergence at the fine-tuning stage.

*D. Testing and calculation of results*
After training each DBN, the accuracy of each $DBN_i$ is determined. Each DBN is associated with a weight $w_i$
where

$$w_i = 1 - \frac{Number\ of\ misclassified\ images\ in\ training\ set}{Total\ number\ of\ images\ in\ the\ training\ set}$$

Thus, higher weights are assigned to the DBN with higher accuracy. The final classification result for a test image is obtained by voting using the weights of each DBN. The algorithm for calculating the weighted votes for each image is given below.

**Algorithm1**: Weighted voting
Input: Number of classes, n
Output: Predicted value, p
1.     Initialize t = zero matrix (size: n x 1)
2.     for i = 1,….n
3.         for j = 1,2…….16
4.             if i is equal to predicted class
5.                 t(i) = t(i) + $w_i$
6.     p = max(t)

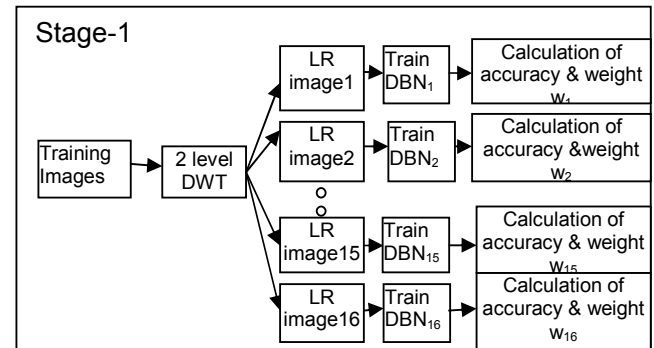

*Figure 1* Preprocessing of Training dataset and training of DBNs

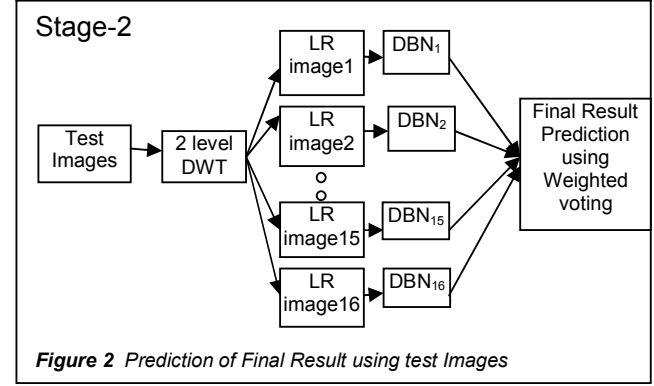

*Figure 2* Prediction of Final Result using test Images

The index i with maximum value in t is taken as the predicted value. The above steps are visualized in Fig. 1 and Fig.2.

*Experimental Results and Discussion:* In order to evaluate the proposed approach, experiments are performed on two standard datasets. The number of hidden units and learning rate of DBNs are decided empirically.

1. COIL-20 Dataset

Columbia Object Image Library (COIL-20) [6] consists of gray-scale images of 20 objects, as shown in Figure 3.

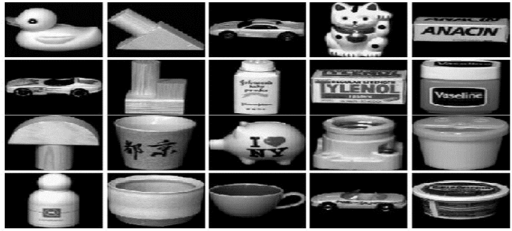

**Figure 3** *20 Objects of COIL-20 Dataset*

Size of each image is 128 x 128. Each image is down-sampled to a size of 64 x 64 and 2 level DWT is taken subsequently. This results in sixteen images of size 16 x 16. Each of these images is reshaped to 1 x 256 and passed on as inputs to train their corresponding DBNs.

The results for this dataset are compared with that of the wavelet transform based method adopted in [2], in which 5 objects are chosen at random. These 5 categories are evaluated at 75/25 hold out ratio (training images/test images) to compare the results obtained using the approach proposed in this letter and those specified in [2] as shown in Table 1.

Architecture of Each DBN: 10 hidden units and 5 output units
Architecture of ANN used in [2]: 82 hidden units and 5 output units

**Table 1** *(Comparison of the results mentioned in [2] and results obtained using the approach in this letter)*

| Object ID | Approach from [2] | | Our Approach | |
|---|---|---|---|---|
| | Training (%) | Testing (%) | Training (%) | Testing (%) |
| 1 | 84.38 | 82.31 | 100 | 94.44 |
| 6 | 79.12 | 77.62 | 100 | 94.44 |
| 8 | 81.23 | 79.89 | 100 | 100 |
| 11 | 80.86 | 79.22 | 100 | 94.44 |
| 19 | 83.28 | 81.63 | 100 | 94.44 |

Results on the whole dataset are compared with that of methods adopted in [7] and [8] for 70/30 hold out ratio (training images/test images) in Table-2. For whole dataset,
Architecture of DBN: [40,20,20]
Time taken to train each DBN: 3.55 sec on Intel Core i5-4200U CPU @1.60 GHz

**Table 2** *Results on COIL-20 dataset*

| | Accuracy (%) |
|---|---|
| MPS [13] | 99.74 |
| S-LE [12] | 99.72 |
| Approach in this letter | 99.72 |

2. USPS Dataset

The USPS data set [9] consists of images of handwritten digits, collected from mail envelopes. Each image is of size 16 x 16. The dataset is divided into a training set (consisting of 7291 images) and a testing set (consisting of 2007 images). Some random sample images from this dataset are shown in Fig. 4.

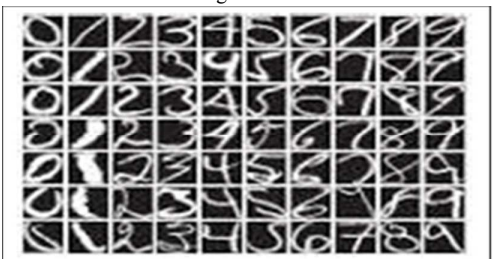

**Figure 4** *Samples from USPS Dataset*

**Table 3** *(Results on USPS dataset using the approach in this letter; CPU: Intel Core i5, 4200 U @1.6 GHz)*

| Architecture | Epochs | Error (%) | Computation Time per DBN |
|---|---|---|---|
| 20 hidden units + 10 output units | 400 | 7.3 | 44.56 sec |
| 50 hidden units + 10 output units | 300 | 6.95 | 37.02 sec |
| 40 hidden units + 20 hidden units + 10 output units | 300 | 6.3 | 47.8 sec |

Results for other DBN based methods [10] are given in Table 4

**Table 4** *(Results on USPS dataset mentioned in [10]; CPU: AMD Opteron processor 8435, 2.6 GHz)*

| Architecture | Error (%) | CPU Time |
|---|---|---|
| DBN | 5.15 | 21.43 H |
| MN DBN(20) | 4.7 | 8.9 H |
| MN DBN (5) | 5.75 | 5.5H |
| SMN DBN (5) | 7.7 | 5.6 H |
| SDBN | 5.1 | 11.9 H |

Therefore, from the results in Table 3 and Table 4, it can be concluded that the approach in this letter can give faster results compared to other DBN based methods listed in [10], albeit with a little trade off in accuracy. Human error rate for USPS dataset is 2.5%.

*Conclusion:* Results on two datasets: COIL-20 and USPS show that this approach can achieve competent results with much less computation time because of reduction in size of input data to DBNs and less number of weight parameters. Since the architecture of DBNs trained using this approach is very small as compared to that of DBNs trained on raw pixels, this system can also be extended to image recognition problems with high resolution. If all the DBNs can be executed in parallel, this approach can be used to train DBNs and test them on a moderately powerful processor (such as the one used in this paper) in a matter of few minutes even for large datasets, with competent accuracy.

Saurabh Sihag and Pranab Kr Dutta (*Department of Electrical Engineering, IIT Kharagpur, Kharagpur-721302, India* )

E-mail: pkd@ee.iitkgp.ernet.in